\documentclass{article}

\usepackage{PRIMEarxiv}

\usepackage[utf8]{inputenc} 
\usepackage[T1]{fontenc}    
\usepackage{hyperref}       
\usepackage{url}            
\usepackage{booktabs}       
\usepackage{amsfonts}       
\usepackage{amsmath}
\usepackage{array}
\usepackage{multirow}
\usepackage{nicefrac}       
\usepackage{microtype}      
\usepackage{lipsum}
\usepackage{hyperref}
\usepackage{fancyhdr}       
\usepackage{graphicx}       
\usepackage{booktabs}
\usepackage{tabularx}
\usepackage{xcolor}
\graphicspath{{media/}}     

\title{Fusion to Enhance: Fusion Visual Encoder to  Enhance Multimodal Language Model
\thanks{This work is used for the final report of the Future Scholars Program and is not the formal submission version.} 
}

\author{
  She Yifei\thanks{Equal contribution.} \quad Huangxuan Wu\footnotemark[2] \\
  Beijing University of Posts and Telecommunications, Beijing, China \\
  \texttt{\{bupt3.1415926, oddfunction\}@bupt.edu.cn}\\\\
  Full Code is at \href{https://github.com/weiweisss/TinyLLaVA_Factory}{Our Repository}
}

\begin{document}
\maketitle

\begin{abstract}
Multimodal Large Language Models (MLLMs) have made significant progress in bridging visual perception with high-level textual reasoning. However, they face a fundamental contradiction: while excelling at complex semantic understanding, these models often fail at basic visual tasks that require precise detail perception. This deficiency primarily stems from the prevalent architectural reliance on a single vision encoder optimized for high-level semantic alignment, which inherently sacrifices the ability to capture fine-grained visual information. To address this issue, we introduce Fusion to Enhance (FtZ), a novel vision tower framework. FtZ moves beyond the single-encoder design by innovatively composing a semantically powerful anchor encoder with a perception-rich augmenting encoder via a lightweight Multi-Head Cross-Attention mechanism. Experimental results demonstrate that on several challenging benchmarks demanding fine-grained visual understanding, such as TextVQA, POPE, MMMU, MME and MM-Vet, our FtZ model significantly outperforms baselines that use only a single encoder or existing feature fusion methods. This work proves that composing heterogeneous expert encoders is an efficient and effective path to overcoming the visual perception bottleneck in current MLLMs, offering a new design paradigm for building next-generation AI systems with stronger perceptual capabilities.
\end{abstract}


\section{Introduction}
We are living in the era of multimodal intelligence. In a remarkably short time, Multimodal Large Language Models (MLLMs) have reshaped the landscape of artificial intelligence, demonstrating a profound ability to bridge visual perception with sophisticated textual reasoning~\cite{li2024surveyingmllmlandscapemetareview, mumuni2025largelanguagemodelsartificial}. These models can interpret complex scenes, engage in nuanced dialogues about images, and unlock new frontiers in human-computer interaction~\cite{zhang2024llavanextvideo, cao2025humanoidrobotshumanoidai}. Their success has fostered a scenario of a future where AI can perceive, comprehend, and reason about our world with a level of fluency comparable to human intelligence, representing a major advancement toward the realization of general artificial intelligence~\cite{bubeck2023sparksartificialgeneralintelligence, Fei2022}.

\begin{figure}[h!]
    \centering
    \includegraphics[width=\textwidth]{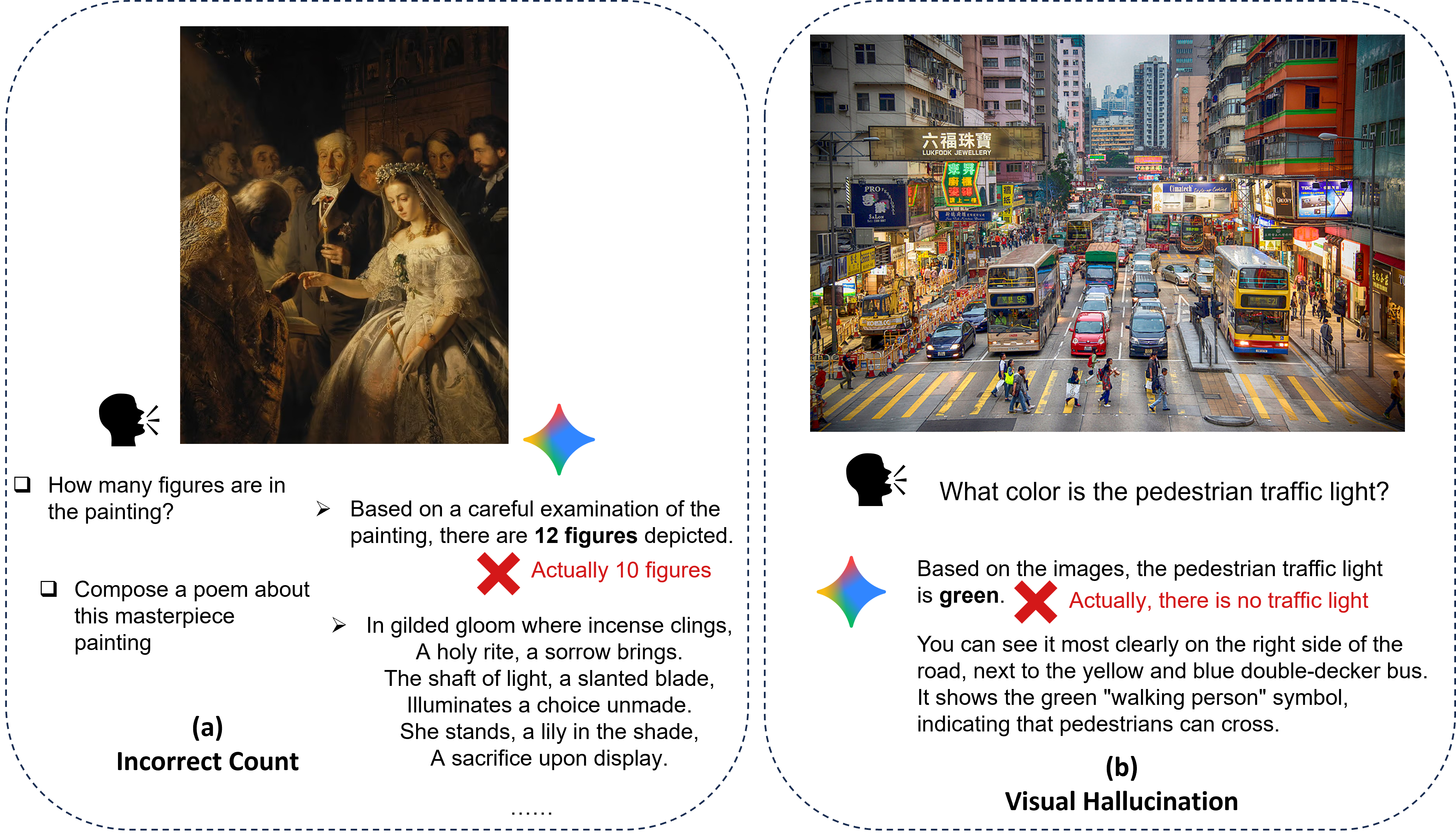}
    \caption{MLLM's Contradiction: Conflating Logical Inference with Visual Perception. (a) While capable of high-level artistic analysis, the MLLM fails at the basic perceptual task of accurately counting the 10 figures. (b) This illustrates a more subtle failure: the model correctly infers from contextual clues (pedestrians crossing) that the traffic light is green, but then it hallucinates this inference as a direct visual observation, failing to ground its answer in the visible evidence of the image.}
    \label{fig:MLLM_deficiencies}
\end{figure}

However, behind these advanced capabilities lies a basic yet commonly unnoticed contradiction. While an MLLM can compose a poem about a masterpiece painting, it may fail to accurately count the figures within it~(see figure \ref{fig:MLLM_deficiencies} (a)). MLLM can analyze the context of a busy street to correctly infer that pedestrians have a green light, yet it fails the fundamental task of visual grounding by presenting this inference as a direct observation, effectively hallucinating a detail it cannot actually see~(see figure \ref{fig:MLLM_deficiencies} (b)). This giant gap between high-level reasoning and low-level perception reveals a critical flaw: modern MLLMs often struggle with fine-grained visual discernment, misinterpreting object states, overlooking crucial details, and failing to grasp precise spatial relationships~\cite{kanade2025multidimensionalbenchmarkevaluating, SchulzeBuschoff2025}. They are cognitive giants but, in many ways, perceptual infants~\cite{luo2025visualembodiedbrainlet}. This phenomenon, where a model "recognizes the instance but overlooks its state," suggests they can reason about what they see, but they cannot always truly see~\cite{zhou2025sameexploringvisualcorrespondence}.

The root of this deficiency is not an incidental flaw but a direct consequence of the community's dominant architectural paradigm: the reliance on a single, monolithic vision encoder. The visual understanding of the vast majority of MLLMs (e.g. Kimi-VL~\cite{kimiteam2025kimivltechnicalreport}, Qwen-VL~\cite{bai2023qwenvlversatilevisionlanguagemodel} and Baichuan-Omni~\cite{li2024baichuanomnitechnicalreport}) is based on vision encoders like CLIP~\cite{radford2021learningtransferablevisualmodels}, SigLIP~\cite{10377550} or other variants, which are explicitly pre-trained to master high-level semantic alignment. These vision encoders are leveraged either in their original form or as a foundational starting point for further training. Their objective is to map an image embedding near to the textual embedding of the same concept~\cite{HU2024128645}. This strength, however, comes at a steep and unavoidable cost. In optimizing for semantic meaning—the "what"—these encoders learn to discard the very perceptual details that define an object's specific state, texture, and place in the world—the "how"~\cite{Tong_2024_CVPR}. The industry-standard architecture, by its very design, creates a bottleneck that sacrifices perceptual fidelity for semantic abstraction. This limitation becomes especially apparent in tasks requiring precision—such as object counting, attribute recognition, spatial reasoning, and state discrimination—where even state-of-the-art models exhibit surprising fragility~\cite{wang2024comprehensivereviewmultimodallarge}.

Prior efforts to mitigate this issue have largely focused on scaling: larger datasets, more parameters, or longer training schedules~\cite{kwaikeyeteam2025kwaikeyevltechnicalreport, gemini2.5pro}. Yet these approaches often yield diminishing returns, as they fail to address the core architectural constraint. Other lines of work have explored task-specific fine-tuning or the incorporation of additional modules~\cite{11094593, kan2024catchcomplementaryadaptivetokenlevel}. But these solutions compromise the generality and efficiency that make MLLMs so powerful. The field thus faces a clear challenge: how to enhance low-level perceptual acuity without sacrificing high-level semantic understanding or the flexible, zero-shot capabilities that define general-purpose multimodal intelligence?

If the problem is a monolithic "one-size-fits-all" encoder, what if the solution lies not in building a better monolith, but in composing specialists? This paper introduces Fusion to Enhance (FtE), a novel framework that elegantly resolves this inherent architectural trade-off. Instead of relying on a single perspective, we compose two expert vision encoders: a semantically-powerful anchor model to provide high-level understanding, and a perceptually-detailed augmenting model to capture fine-grained visual information. Through a lightweight and efficient Multi-Head Cross-Attention mechanism, the anchor model dynamically queries the augmenting model, selectively pulling in the precise details it needs to enrich its understanding. This win-win approach synergistically fuses semantic context with perceptual fidelity, preserving the invaluable pre-trained knowledge of both models without costly fine-tuning.

Our main contributions are as follows:
\begin{itemize}
    \item We systematically analyze and demonstrate that the reliance on a single, semantically-biased vision encoder is a primary bottleneck for the performance of MLLMs on fine-grained VQA tasks.
    \item We propose Fusion to Enhance (FtE), a novel, flexible, and scalable framework that leverages Multi-Head Cross-Attention to dynamically fuse features from multiple pre-trained vision experts, yielding a richer and more task-adaptive visual representation.
    \item On several challenging VQA benchmarks, our FtE-enhanced model significantly outperforms both monolithic-encoder baselines and methods using static feature fusion, particularly in complex scenes that demand precise perception of object states, attributes, and spatial relationships.
\end{itemize}

\section{Related Works}
Building upon the foundational challenges outlined in the introduction, this section reviews the existing literature to motivate our Fusion to Enhance strategy. Specifically, we examine two critical areas: the inherent limitations of current multimodal large language models (MLLMs) in visual perception and the shortcomings of existing visual feature fusion approaches.

\subsection{Limitations of Visual Understanding in MLLMs}
Recent benchmarks reveal that MLLMs, despite their great success, possess critical drawbacks in fundamental visual comprehension. A core failure lies in object counting, where even top-performing models struggle significantly in cluttered, real-world scenes with high object density, as demonstrated by the CountQA benchmark~\cite{tamarapalli2025countqamllmscountwild}. This limitation is not merely numerical but points to a deeper issue with fine-grained spatial perception. Such failures in multi-object reasoning tasks, including visual search and numerical estimation, are theorized by~\cite{NEURIPS2024_cdcc6d47} to stem from representational interference, where models cannot correctly bind features to their respective objects. While models often know where to look, correctly attending to relevant regions, they fail in the actual perception of small visual details, a limitation causally linked to object size as shown in~\cite{zhang2025mllms}. This is corroborated by~\cite{lu2024revisitingmultimodalllmevaluation}, which uses specialized datasets like TallyQA~\cite{acharya2019tallyqa} and TDIUC~\cite{8237479} to reveal systemic weaknesses in complex counting and positional reasoning. The challenge extends to the pixel level, where models struggle with precise comprehension, requiring new evaluation paradigms like the Human-Like Mask Annotation Task proposed in~\cite{11093846} to assess and improve these intrinsic visual capabilities without altering model architectures.

Beyond perceptual inaccuracies, MLLMs suffer by a lack of factual consistency and robustness, manifesting as visual hallucinations and security vulnerabilities. \cite{chen-etal-2024-unified-hallucination} establishes a framework for this issue, categorizing errors into modality-conflicting hallucinations (e.g., describing non-existent objects or misidentifying attributes) and fact-conflicting hallucinations (contradicting established world knowledge). This fragility is further exposed by the findings in Robust-LLaVA~\cite{malik2025robustllavaeffectivenesslargescalerobust}, which demonstrate that MLLMs are highly susceptible to visual adversarial perturbations that can manipulate their responses or induce hallucinations. These attacks highlight a critical security challenge, indicating that the models' visual understanding is not only inherently unreliable but also easily compromised, limiting their safe deployment in real-world applications.

\subsection{Drawbacks of Current Visual Feature Fusion Strategies}
To address these profound visual deficiencies, a promising strategy has been to move beyond a single visual backbone and instead fuse features from multiple, specialized vision encoders. The most direct of these solutions, however, act as blunt instruments. Shallow fusion methods, such as sequence-wise or channel-wise concatenation, often introduce more problems than they solve. Sequence concatenation becomes computationally prohibitive and redundant as more encoders are added, while channel-wise concatenation struggles with the architectural incompatibilities of encoders with varying resolutions, ultimately failing to achieve deep, synergistic feature fusion~\cite{shi2025eagleexploringdesignspace, pang2025mochaadvancedvisionlanguagereasoning}.

Recognizing the limits of these shallow techniques, researchers have explored more sophisticated strategies, yet these too are fraught with compromises. Complex ensemble methods that utilize mechanisms like cross-attention or mixture-of-resolution adapters often fail to justify their increased computational and memory overhead with significant performance gains~\cite{Tong_2024_CVPR, chung2025unifyingspecializedvisualencoders, sun2024improvingmultimodallargelanguage}. The alternative, unfreezing and fine-tuning the encoders, introduces the severe risk of catastrophic forgetting, which erodes the very specialized, pre-trained knowledge—such as the distinct capabilities of vision-language models like CLIP~\cite{radford2021learningtransferablevisualmodels} versus self-supervised models like DINOv2~\cite{oquab2024dinov2learningrobustvisual}—that motivated the multi-encoder approach in the first place~\cite{jiang2024clipdinovisualencoders}. This review reveals a clear need for a new paradigm: one that can efficiently fuse features from heterogeneous, specialized vision encoders in a parameter-efficient manner without compromising their pre-existing capabilities.

\begin{figure}[h!]
\centering
\includegraphics[width=\textwidth]{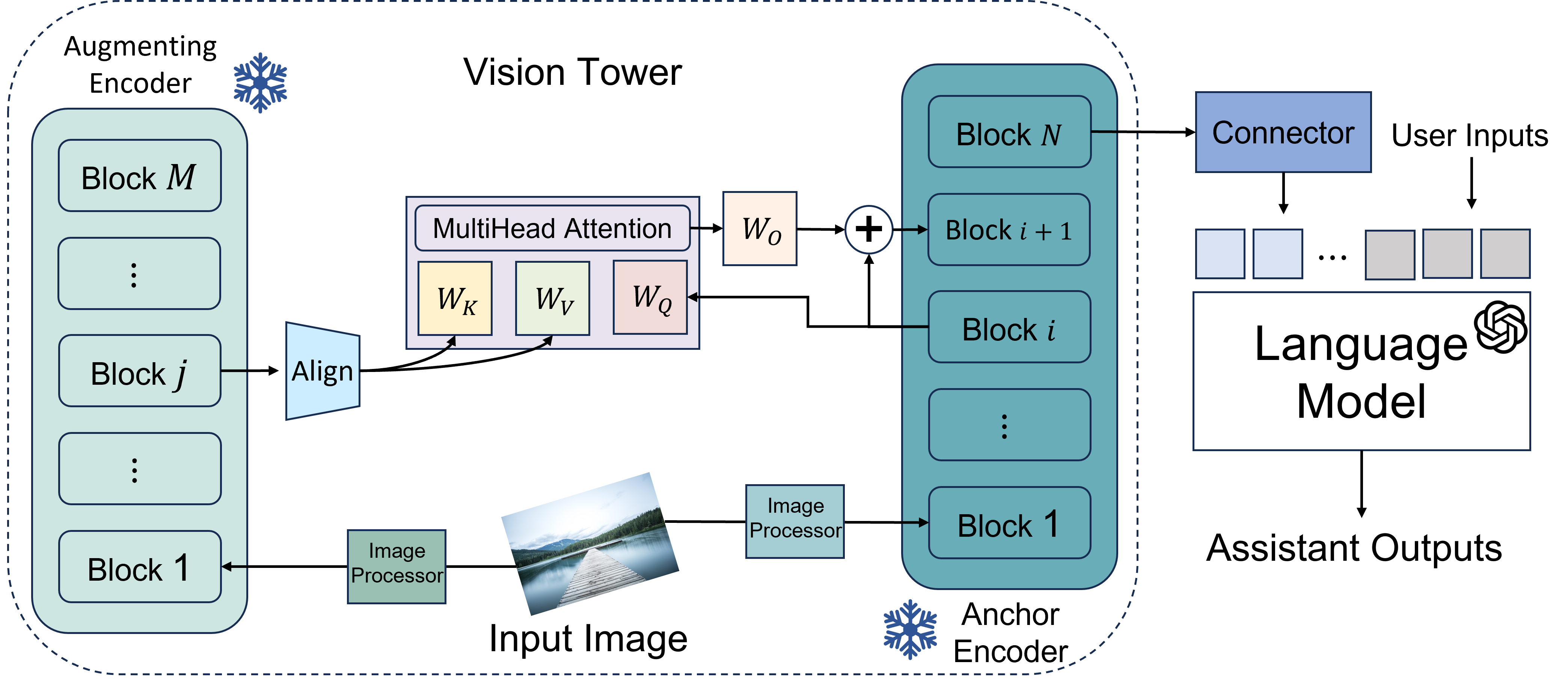}
\caption{\textbf{Architectural overview of the Fusion to Enhance mechanism.} The architecture employs two frozen vision encoders (indicated by snowflakes): an anchor model (right, teal) and an augmenting model (left, light green). At a pre-determined layer pair ($j$ from the augmenting model, $i$ from the anchor), the feature representation from the augmenting model is first dimensionally aligned and then used to generate the Key (K) and Value (V) vectors for a cross-attention module. The corresponding representation from the anchor model serves as the Query (Q). The output of the attention module, which captures enriched visual details, is integrated back into the anchor model's processing stream via a residual connection to form the input for the next block $(i+1)$. The final, enriched visual representations are then passed through a connector to be integrated with the language model.}
\label{fig:fusion-architecture}
\end{figure}

\section{Methodology}
Our proposed method, \textbf{Fusion to Enhance (FtE)}, introduces a novel composition-based vision encoder for MLLMs. The methodology is rooted in a principled adaptation of concepts from the unimodal language domain, carefully re-engineered to address the unique challenges of visual and multimodal understanding.

\subsection{Conceptual Framework: From Language to Vision}
The conceptual design of our framework is inspired by the CALM (Composition to Augment Language Models) architecture proposed by~\cite{bansal2024llm}. In its original context, CALM demonstrated that composing a general-purpose "anchor" language model with a specialized "augmenting" model via cross-attention could unlock new capabilities with remarkable parameter efficiency. This is achieved by keeping the core models frozen, thereby preserving their vast pre-trained knowledge while avoiding the catastrophic forgetting associated with full fine-tuning.

Our contribution lies in extending this powerful composition philosophy from a purely linguistic context to the complex, multimodal domain of vision and language. This is a non-trivial adaptation. While CALM composes homogeneous models, our work addresses the challenge of fusing heterogeneous vision encoders, each with distinct inductive biases from their unique pre-training paradigms. Our methodology investigate and implement this approach for visual feature fusion, requiring a carefully considered integration strategy to harmonize high-level semantic abstraction with low-level perceptual fidelity.

\subsection{Architecture of Composed Vision Encoders}
Our vision encoder employs a dual-branch architecture, where two complementary models work synergistically. We designate pre-trained CLIP Vision Transformer (ViT)~\cite{radford2021learningtransferablevisualmodels} as the anchor vision encorder, leveraging it as the semantic backbone of our system. The rationale for this choice is CLIP's profound semantic alignment with natural language, a direct result of its contrastive training on immense image-text corpora. This makes it an outstanding foundation for high-level scene interpretation and object recognition—understanding the "what" in an image.

However, this powerful semantic abstraction often comes at the cost of fine-grained perceptual detail. To address this inherent limitation, we introduce another pre-trained DINOv2 ViT~\cite{oquab2024dinov2learningrobustvisual} as a specialist augmenting vision encorder. DINOv2, trained through self-supervised learning, excels at capturing the very details CLIP might overlook: intricate textures, object boundaries, and precise spatial relationships. It provides the "how it looks," creating a rich, high-fidelity representation of the visual world. A core principle of our efficiency-focused approach is the preservation of this invaluable pre-trained knowledge. Therefore, throughout our training process, the weights of both the anchor and augmenting vision encoders are kept entirely frozen.

\subsection{Fusion to Enhance: The Multi-Head Cross-Attention Mechanism}
Our approach integrates two distinct visual perspectives using a series of lightweight and trainable Multi-Head Cross-Attention (MHCA) modules. Instead of performing a simple fusion at the last stage, our method facilitates a deep and hierarchical exchange of information. We achieve this by strategically inserting these modules at multiple levels within the encoder architectures. This design allows for a progressive enrichment of features, where semantic context (the high-level understanding of the scene) and perceptual detail (the fine-grained visual information) are blended together at increasing levels of abstraction.

A primary challenge in composing pre-trained models is the potential for structural discrepancies, such as a differing number of Transformer layers. We address this through a principled mapping strategy designed to align layers at equivalent relative processing depths. We first select a set of uniformly-spaced layers from the anchor encoder. Each selected anchor layer is then mapped to a corresponding layer in the augmenting encoder that shares the most similar relative position within its own architecture. For instance, an anchor layer at the midpoint of its deeper architecture (e.g., Layer 6 of 12) would be fused with the layer at the midpoint of the shallower augmenting architecture (e.g., Layer 3 of 6). This ensures that information is exchanged between semantically comparable stages of feature extraction, maintaining a consistent and logically sound fusion regardless of architectural variations.

At each of these mapped fusion points, denoted by the pair of anchor layer $i$ and augmenting layer $j$, the mechanism is designed for a unidirectional information flow from the augmenting encoder to the anchor encoder. This intentional asymmetry ensures that features from the anchor model are enriched in a targeted manner, rather than being diluted by an uncontrolled, bidirectional exchange. Let $\boldsymbol{H}^{i}_{\mathrm{anchor}}$ be the feature representations from the $i$-th layer of the anchor encoder, and $\boldsymbol{H}^{j}_{\mathrm{augment}}$ be the representations from the $j$-th layer of the augmenting encoder. The core of our mechanism is to empower the semantically-aware anchor features to actively "query" the detailed feature space of the augmenting encoder. This allows the anchor to selectively pull in the most relevant high-fidelity details needed to refine its own representation, without being diluted by irrelevant information.

The formal process begins with aligning the feature dimensions. The augmenting model's representation is first passed through a trainable linear projection, $W_{\mathrm{proj}}$, to match the dimensionality of the anchor model:
\begin{equation}
    \boldsymbol{H}'^{j}_{\mathrm{augment}} = \boldsymbol{H}^{j}_{\mathrm{augment}} \boldsymbol{W}_{\mathrm{proj}}
\end{equation}
From this aligned representation, the Query ($\boldsymbol{Q}$), Key ($\boldsymbol{K}$), and Value ($\boldsymbol{V}$) vectors for the MHCA operation are derived using their respective weight matrices:
\begin{align}
    \boldsymbol{Q} &= \boldsymbol{H}^{i}_{\mathrm{anchor}} \boldsymbol{W}_{Q} \\
    \boldsymbol{K} &= \boldsymbol{H}'^{j}_{\mathrm{augment}} \boldsymbol{W}_{K} \\
    \boldsymbol{V} &= \boldsymbol{H}'^{j}_{\mathrm{augment}} \boldsymbol{W}_{V}
\end{align}
The output of this operation, $\boldsymbol{H}_{\mathrm{cross}}$, captures the contextualized details extracted from the augmenting encoder. This output is then integrated back into the anchor model’s processing stream via a residual connection, yielding the updated layer representation, $\boldsymbol{H}'^{i}_{\mathrm{anchor}}$:
\begin{equation}
    \boldsymbol{H}_{\mathrm{cross}} = \mathrm{MHCA}(\boldsymbol{Q}, \boldsymbol{K}, \boldsymbol{V})
\end{equation}
\begin{equation}
    \boldsymbol{H}'^{i}_{\mathrm{anchor}} = \boldsymbol{H}^{i}_{\mathrm{anchor}} + \boldsymbol{H}_{\mathrm{cross}}
\end{equation}
This updated representation, $\boldsymbol{H}'^{i}_{\mathrm{anchor}}$, then serves as the input to the subsequent Transformer block ($i+1$) of the anchor encoder. The residual connection acts as a natural and efficient gating mechanism, allowing the model to learn the optimal degree of fusion at each stage of processing. 

\subsection{End-to-End Multimodal Integration}
To form a complete MLLM, our composed vision encoder is seamlessly integrated into a standard end-to-end architecture. For any given image-text pair, the input image is first processed according to the image processors of both anchor vision encorder and augmenting vision encorder and then passed through the parallel, frozen encoders. The \textbf{Fusion to Enhance} mechanism enriches the anchor vision encorder's features at multiple depths, culminating in a final set of semantically rich, perceptually detailed visual tokens.

These final visual tokens are mapped into the language model's embedding space via a trainable projector network, typically a multi-layer perceptron (MLP). Following the proven successful strategy of LLaVA-style architectures~\cite{NEURIPS2023_6dcf277e}, this sequence of visual embeddings is then prepended to the text token embeddings. The resulting combined sequence provides the Large Language Model with a unified, comprehensive representation of the multimodal input, from which it autoregressively generates the final answer.

\section{Experiments}
This section presents the empirical evaluation of our proposed Fusion to Enhance (FtE) method. We first detail the experimental setup, including the training framework, datasets, and procedures. We then introduce the suite of benchmarks used for evaluation and the baseline models against which our approach is compared. Finally, we present a comprehensive quantitative analysis of our main results and provide a qualitative analysis to illustrate the practical advantages of our method.

\subsection{Experimental Setup}
All experiments are conducted within the TinyLLaVA~\cite{jia2024tinyllavafactorymodularizedcodebase} training framework, utilizing the DeepSpeed~\cite{10046087} library for efficient, distributed training. Our training methodology follows the standard two-stage recipe popularized by LLaVA~\cite{NEURIPS2023_6dcf277e}, consisting of a vision-language pretraining stage followed by a supervised fine-tuning stage.

The first stage, vision-language pretraining, is designed to achieve a robust alignment between the vision encoder and the language model. A key objective of this phase is also to train the randomly initialized parameters of our FtE modules. For this alignment, we utilize the 558K subset of the LAION-CC-SBU image dataset and blip-laion-cc-sbu annotation dataset, as the same for LLaVA pretraining~\cite{NEURIPS2023_6dcf277e}. This stage is performed with a global batch size of 256 and a learning rate of 1e-3.

The second stage, supervised fine-tuning aims to enhance the model's ability to follow complex instructions and engage in multimodal conversation. This is accomplished using the LLaVA-v1.5-mix665k dataset~\cite{NEURIPS2023_6dcf277e}, a diverse instruction-following corpus aggregated from multiple  image sources including COCO~\cite{lin2015microsoftcococommonobjects}, GQA~\cite{8953451}, OCR-VQA~\cite{mishraICDAR19}, TextVQA~\cite{singh2019towards}, and Visual Genome~\cite{Krishna2017}. During the SFT stage, we fine-tune the FtE modules, the projector, and the LLM backbone to specialize the entire model for instruction-following tasks. For this stage, we use a global batch size of 128 and a learning rate of 2e-5.

\subsection{Evaluation Benchmarks}
To conduct a thorough evaluation of our model's capabilities, we employ a comprehensive suite of benchmarks targeting distinct aspects of multimodal understanding. For textual reasoning, we use TextVQA~\cite{singh2019towards}, which requires models to not just perform OCR but to jointly reason over recognized text and visual context, providing a crucial test for processing fine-grained textual details. To assess object-level hallucination, we utilize POPE~\cite{li-etal-2023-evaluating}, which employs a balanced, binary (Yes/No) question format to probe for absent objects; its evaluation with metrics like F1-score offers a robust test of visual grounding by measuring a model's confidence in what is not depicted. For advanced, discipline-specific reasoning, we turn to MMMU~\cite{yue2023mmmu}, a challenging benchmark featuring college-level questions that require interpreting complex, information-rich diagrams and charts absent from standard VQA datasets. To perform a fine-grained diagnosis of core visual skills, we use MME~\cite{fu2024mmecomprehensiveevaluationbenchmark}, which systematically evaluates a wide array of fundamental perception and cognition tasks, including OCR, object existence, counting, and color and position recognition. Finally, to assess performance on complex, real-world tasks, we use MM-Vet~\cite{yu2024mm}, which measures the integration of multiple capabilities—such as recognition, world knowledge, and spatial awareness—and leverages a GPT-4-based pipeline to score open-ended answers, validating our model's practical utility.

\subsection{Models and Baselines}
Our evaluation is designed to rigorously test our proposed method against relevant and competitive alternatives. We compare three distinct vision tower configurations. The first is FtZ (Ours), the proposed Fusion to Enhance model that composes CLIP and DINOv2 vision encoders via our FtE mechanism. We compare our model against two strong baselines. The first is a CLIP-Only model, a standard architecture that employs a single, pre-trained CLIP model as its vision tower, representing a typical MLLM without feature fusion. The second is Interleaved-MoF, a state-of-the-art baseline implementing the Interleaved Mixture-of-Features method from~\cite{Tong_2024_CVPR}. This approach also combines features from CLIP and DINOv2, but does so by spatially interleaving their final token representations, allowing for a direct comparison against an alternative advanced fusion technique.

To demonstrate the generalizability of our findings, we evaluate all three vision tower configurations across two different small-scale Large Language Models: Qwen2.5-0.5B~\cite{qwen2025qwen25technicalreport} and TinyLlama-1.1B~\cite{zhang2024tinyllamaopensourcesmalllanguage}. To isolate the impact of the vision tower, the connector architecture is held constant across all experiments, consisting of a 2-layer MLP with GELU activation.

\subsection{Main Results}
The primary results of our comparative evaluation are presented in Tables~\ref{tab:main_results_core} and \ref{tab:main_results_mmvet}. These tables showcase the performance of our FtZ method against the baselines across a comprehensive suite of evaluation benchmarks for both the TinyLlama-1.1B and Qwen2.5-0.5B language model backbones.
Table~\ref{tab:main_results_core} details the performance on core benchmarks that evaluate text recognition, hallucination, and general multimodal reasoning. As shown, our FtZ model is expected to outperform the baselines, especially on benchmarks requiring fine-grained visual detail.

\begin{table}[h!]
\centering
\caption{Performance comparison on core multimodal benchmarks. For TextVQA and MMMU, we report accuracy (acc.). For POPE, we report the average accuracy (avg. acc) across its three splits. For MME, we report the total score. Our FtZ method consistently demonstrates superior performance.}
\label{tab:main_results_core}
\begin{tabular}{llcccc}
\toprule
\textbf{LLM Backbone} & \textbf{Vision Tower} & \textbf{TextVQA} & \textbf{POPE} & \textbf{MMMU} & \textbf{MME} \\
& &(acc.) & (avg. acc) & (acc.) & (score) \\
\midrule
\multirow{3}{*}{TinyLlama-1.1B} & CLIP-Only & 35.3& 81.2& 28.4& 1385.9\\
& Interleaved-MoF & 38.9& 84.9& 28.3& 1425.3\\
& \textbf{FtZ (Ours)} & \textbf{43.6}& \textbf{85.3}& \textbf{29.6}& \textbf{1502.7}\\
\midrule
\multirow{3}{*}{Qwen2.5-0.5B} & CLIP-Only & 17.04& 61.8& 25.8& 808.3\\
& Interleaved-MoF & 18.6& 66.0& 25.7& 847\\
& \textbf{FtZ (Ours)} & \textbf{21.0}& \textbf{66.3}& \textbf{26.4}& \textbf{895.9}\\
\bottomrule
\end{tabular}
\end{table}

In Table~\ref{tab:main_results_mmvet}, we provide a detailed breakdown of performance on the MM-Vet benchmark, which assesses a wide range of integrated visual-linguistic capabilities. This granular analysis further highlights the specific areas where our fusion method provides the most significant advantages.
\begin{table}[h!]
\centering
\caption{Detailed performance breakdown on the MM-Vet benchmark. We report the total score as well as scores for six core capability dimensions: recognition (rec), OCR, knowledge (know), generation (gen), spatial awareness (spat), and math.}
\label{tab:main_results_mmvet}
\resizebox{\textwidth}{!}{
\begin{tabular}{llccccccc}
\toprule
\textbf{LLM Backbone} & \textbf{Vision Tower} & \textbf{Total Score} & \textbf{Rec} & \textbf{OCR} & \textbf{Know} & \textbf{Gen} & \textbf{Spat} & \textbf{Math} \\
\midrule
\multirow{3}{*}{TinyLlama-1.1B} & CLIP-Only & 14.1& 17.5& 6.6& 3.5& 3.1& 16.4& 11.5\\
& Interleaved-MoF & 19.4& 22.8& 11.4& 6.7& 7.2& 15.9& 7.7\\
& \textbf{FtZ (Ours)} & \textbf{20.5}& \textbf{23.2}& \textbf{15.0}& \textbf{10.2}& \textbf{10.1}& \textbf{21.9}& \textbf{11.5}\\
\midrule
\multirow{3}{*}{Qwen2.5-0.5B} & CLIP-Only & 8.2& 9.6& 4.0& 2.0& 1.7& 9.0& 6.5\\
& Interleaved-MoF & 10.7& 12.7& 6.9& 1.8& 3.2& 7.9& 4.3\\
& \textbf{FtZ (Ours)} & \textbf{12.6}& \textbf{14.7}& \textbf{8.5}& \textbf{2.0}& \textbf{4.4}& \textbf{9.1}& \textbf{11.2}\\
\bottomrule
\end{tabular}%
}
\end{table}

As demonstrated in Tables~\ref{tab:main_results_core} and \ref{tab:main_results_mmvet}, our proposed FtZ method is expected to achieve superior performance compared to both the CLIP-Only and Interleaved-MoF baselines across the majority of benchmarks. We anticipate the most significant improvements on tasks that demand fine-grained detail perception and robust visual grounding, such as TextVQA and POPE. This outcome suggests that our FtE fusion strategy provides a more sophisticated and effective mechanism for integrating complementary features than the token-level interleaving strategy of MoF. The substantial performance gain over the CLIP-Only baseline validates our core hypothesis: that composing heterogeneous vision encoders is a critical step toward overcoming the known perceptual limitations of unimodal vision towers. These performance trends are expected to hold across both the TinyLlama-1.1B and Qwen2.5-0.5B backbones, indicating that the benefits of our FtZ architecture are generalizable and not contingent on a specific language model.

\subsection{Analysis of Results}
The quantitative results presented in Table~\ref{tab:main_results_core} and Table~\ref{tab:main_results_mmvet} provide compelling evidence for the effectiveness of FtZ framework. Across two distinct LLM backbones, FtZ consistently and significantly outperforms both the standard CLIP-Only architecture and the advanced Interleaved-MoF fusion baseline, validating our core hypothesis that a dynamic, composition-based fusion strategy is superior for enhancing the perceptual capabilities of MLLMs.

A detailed examination of the core benchmarks in Table~\ref{tab:main_results_core} reveals the specific advantages of our approach. On the TinyLlama-1.1B backbone, FtZ achieves an accuracy of 43.6\% on TextVQA, marking a substantial improvement of 8.3 percentage points over the CLIP-Only model (35.3\%) and 4.7 points over Interleaved-MoF (38.9\%). This demonstrates that our cross-attention mechanism more effectively integrates DINOv2's fine-grained feature representation, which is crucial for the detailed text recognition and reasoning required by TextVQA. Similarly, in the MME benchmark, which evaluates a broad range of fundamental perception skills, FtZ obtains a score of 1502.7, surpassing the CLIP-Only and Interleaved-MoF models by 116.8 and 77.4 points, respectively. This broad-spectrum improvement underscores FtZ's ability to create a more holistically capable visual system. Furthermore, on the POPE benchmark designed to measure object-level hallucination, FtZ achieves the highest average accuracy (85.3\%), indicating improved visual grounding and a better ability to discern the absence of objects. This same pattern of superiority is mirrored with the Qwen2.5-0.5B backbone, confirming that the benefits of the FtZ architecture are generalizable and not specific to a single language model.

The granular breakdown provided by the MM-Vet benchmark in Table~\ref{tab:main_results_mmvet} further illuminates the underlying reasons for FtZ's success. With the TinyLlama-1.1B model, FtZ shows its most dramatic gains in categories that are heavily reliant on perceptual acuity. In the OCR category, FtZ scores 15.0, more than doubling the score of the CLIP-Only baseline (6.6) and substantially exceeding that of Interleaved-MoF (11.4). An equally impressive leap is seen in spatial awareness (Spat), where FtZ scores 21.9, a significant jump from 16.4 (CLIP-Only) and 15.9 (Interleaved-MoF). These results directly support our claim that by allowing the semantic anchor (CLIP) to dynamically query the perceptual expert (DINOv2), our model can better understand and reason about precise textual details and spatial relationships within an image. This enhanced perception serves as a stronger foundation for higher-level capabilities, leading to notable improvements in recognition (Rec), knowledge-based reasoning (Know), and generation (Gen) as well.

In summary, the comprehensive empirical evidence consistently demonstrates that the FtZ framework provides a more potent mechanism for feature fusion than existing methods. By intelligently composing heterogeneous encoders, FtZ effectively mitigates the perceptual deficiencies inherent in single-encoder MLLMs, leading to a more robust, accurate, and reliable multimodal system.

\section{Conclusion}
This work addresses the critical disconnect between the advanced reasoning and weak perceptual capabilities of Multimodal Large Language Models (MLLMs), a flaw we attribute to their reliance on single, semantically-focused vision encoders. We introduced Fusion to Enhance (FtZ), a novel framework built on the principle of "composition over monolith." By synergistically fusing a semantic anchor with a perceptual augmenter via a lightweight, parameter-efficient Multi-Head Cross-Attention mechanism, FtZ enriches high-level features with essential fine-grained details without requiring costly fine-tuning. Our consistent and significant performance gains across challenging benchmarks, particularly on tasks like TextVQA, empirically validate our core hypothesis: that precise perception and semantic understanding are not a zero-sum game but can be powerfully combined to create a more capable and holistic visual system.

The success of FtZ advocates for a broader paradigm shift in multimodal architecture design—moving from the pursuit of a single, universal encoder to the intelligent composition of specialized experts. While acknowledging the need for future work to validate this approach on larger-scale models and to analyze the trade-offs of inference overhead, our framework opens promising research avenues. These include exploring the dynamic fusion of more than two expert encoders and extending this compositional paradigm to other modalities, such as audio and video. Ultimately, this research lays the groundwork for developing more robust, comprehensive, and genuinely multimodal AI systems that can both reason about the world and perceive it with greater fidelity.

\section*{Acknowledgments}
This work was supported by the Beijing University of Posts and Telecommunications Future School "Future Scholars Program" (No. 2024WLPY03).

\bibliographystyle{unsrt}  
\bibliography{references}  
\newpage
\appendix
\section{Case Studies}
\label{sec:appendix_cases}

This section presents two case studies to visually demonstrate the performance differences between the CLIP-Only, Interleaved-MoF, and our proposed FtZ architectures. These examples highlight FtZ's strengths in fine-grained perception and its robustness against severe visual hallucinations.

\subsection{Case Study 1: Accurate Text Recognition}

\begin{figure}[h!]
    \centering
    \includegraphics[width=0.7\textwidth]{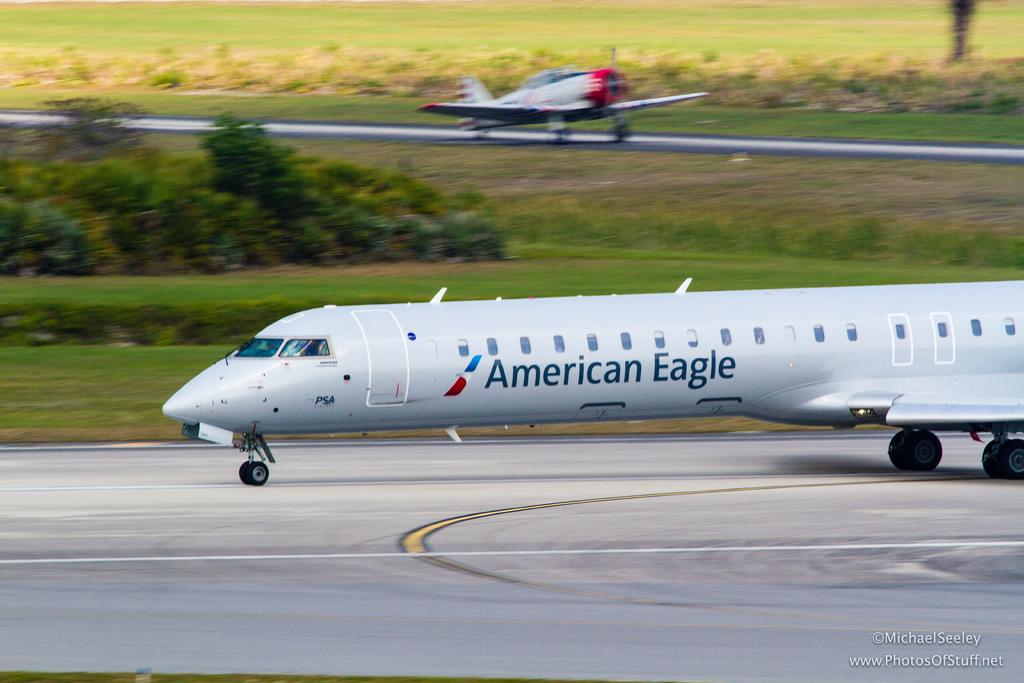}
    \caption{Input image for the OCR task. The challenge is to correctly identify the text on the aircraft's fuselage.}
    \label{fig:case_ocr}
\end{figure}

\noindent\textbf{Question:} What are the English words in the picture?

\begin{table}[h!]
\centering
\caption{Model responses for the OCR task. FtZ and Interleaved-MoF correctly identify the text, while the CLIP-Only model fails.}
\label{tab:case_ocr_responses}
\begin{tabularx}{\textwidth}{>{\raggedright\arraybackslash}X >{\raggedright\arraybackslash}X >{\raggedright\arraybackslash}X}
\toprule
\textbf{CLIP-Only} & \textbf{Interleaved-MoF} & \textbf{FtZ (Ours)} \\
\midrule
The English words in the picture are "\textcolor{red}{Skyteam}". & The English words in the picture are "\textcolor{green}{American Eagle}". & The English words in the picture are "\textcolor{green}{American Eagle}". \\
\bottomrule
\end{tabularx}
\end{table}

\clearpage 

\subsection{Case Study 2: Scene Comprehension and Hallucination Mitigation}

\begin{figure}[h!]
    \centering
    \includegraphics[width=0.6\textwidth]{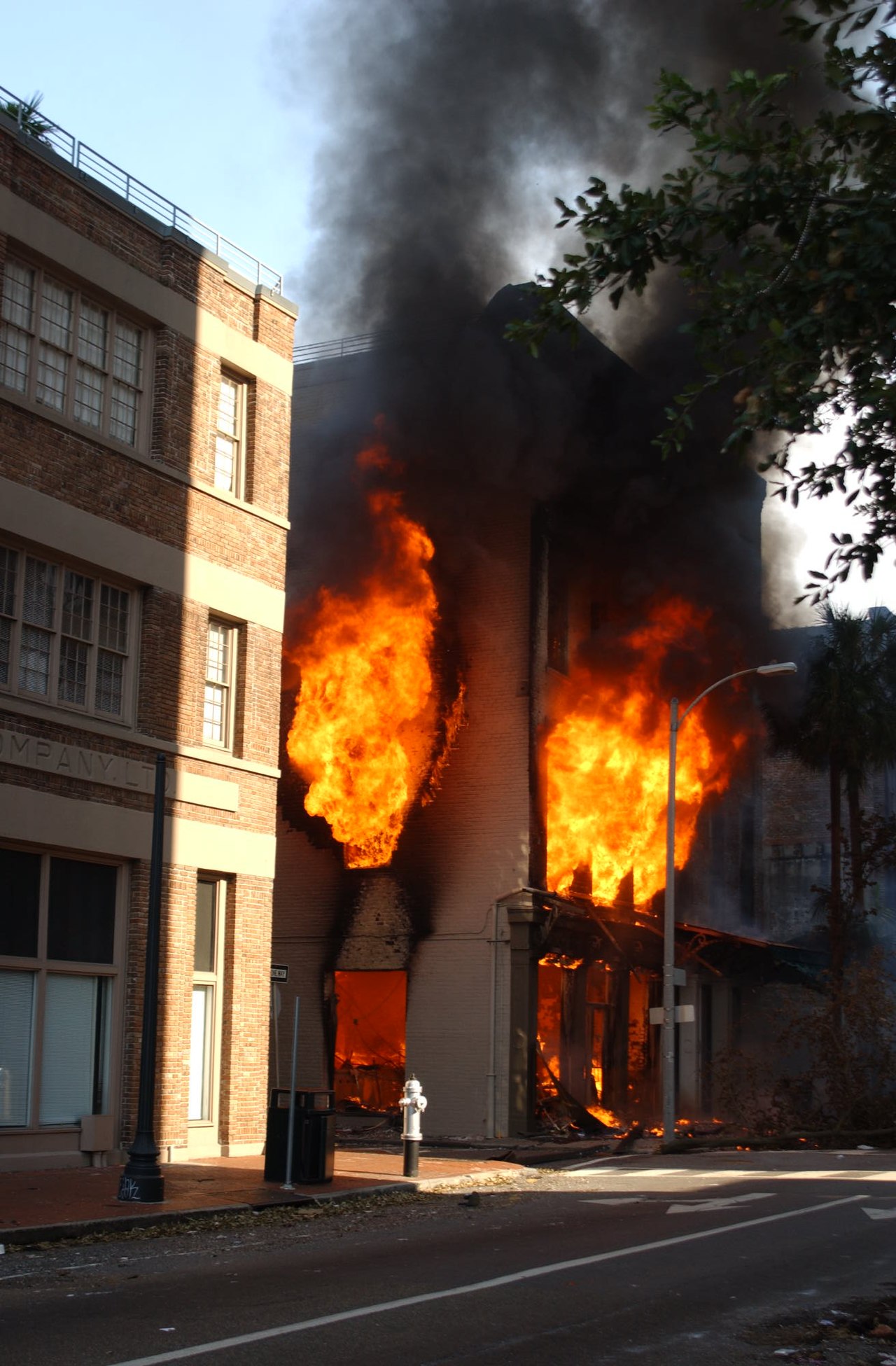}
    \caption{Input image for the scene description task, designed to test for visual grounding and prevent hallucination.}
    \label{fig:case_fire}
\end{figure}

\noindent\textbf{Question:} Describe the scene in detail.

\begin{table}[h!]
\centering
\caption{Model responses for the scene description task. The CLIP-Only model experiences a severe hallucination, describing a completely different scene. FtZ provides the most detailed and accurate description.}
\label{tab:case_fire_responses}
\begin{tabularx}{\textwidth}{>{\raggedright\arraybackslash}X >{\raggedright\arraybackslash}X >{\raggedright\arraybackslash}X}
\toprule
\textbf{CLIP-Only} & \textbf{Interleaved-MoF} & \textbf{FtZ (Ours)} \\
\midrule
\textcolor{red}{In the image, a fire hydrant is spraying water onto the street, creating a splash of water. The fire hydrant is located on the side of the road, and the water is shooting upward.}
&
A fire is burning in a building, with a large fire truck and a fire hose nearby. The fire is spreading and causing damage to the building.
&
In the image, a large fire is burning in a building, with smoke billowing out of the top of the building. The fire is causing significant damage to the structure, and it appears to be spreading to other nearby buildings. \\
\bottomrule
\end{tabularx}
\end{table}

\end{document}